%% file: prcv2020_finalversion.tex
\documentclass[runningheads]{llncs}
\usepackage{graphicx}
\usepackage[namelimits]{amsmath} %
\usepackage{amssymb}             %
\usepackage{amsfonts}            %
\usepackage{mathrsfs}            %
\usepackage{url}

\begin{document}
\title{Confidence-aware Adversarial Learning for Self-supervised Semantic Matching\thanks{This work was supported in part by the Shanghai NSF Grant No.18ZR1425100.}}
\titlerunning{Confidence-aware adversarial learning for Self-supervised Semantic Matching}
%

\author{Shuaiyi Huang\orcidID{0000-0003-0555-2077} \and
	Qiuyue Wang\orcidID{0000-0001-5506-6566} \and
	Xuming He \thanks{Corresponding author.}\orcidID{0000-0003-2150-1237}}

\authorrunning{S. Huang et al.}
%

\institute{
	ShanghaiTech University, Shanghai, China\\
	\email{\{huangshy1,wangqy2,hexm\}@shanghaitech.edu.cn}}
\maketitle              
\begin{abstract}
In this paper, we aim to address the challenging task of semantic matching where matching ambiguity is difficult to resolve even with learned deep features. We tackle this problem by taking into account the confidence in predictions and develop a novel refinement strategy to correct partial matching errors. Specifically, we introduce
a Confidence-Aware Semantic Matching Network (CAMNet)
which instantiates two key ideas of our approach.
First, we propose to estimate a dense confidence map for a matching prediction through self-supervised learning. Second, based on the estimated confidence, we refine initial predictions by propagating reliable matching to the rest of locations on the image plane. In addition, we develop a new hybrid loss in which we integrate a semantic alignment loss with a confidence loss, and an adversarial loss that measures the quality of semantic correspondence. We are the first that exploit confidence during refinement to improve semantic matching accuracy and develop an end-to-end self-supervised adversarial learning procedure for the entire matching network. We evaluate our method on two public benchmarks, on which we achieve top performance over the prior state of the art. We will release our source code at \url{https://github.com/ShuaiyiHuang/CAMNet}.

\end{abstract}
%
%
%

\input{data/intro}

\input{data/related_work}

\input{data/method}

\input{data/loss}

\input{data/experiments}

\input{data/conclusion}
%
%
%
\bibliographystyle{splncs04}
\bibliography{prcv2020}
\end{document}

%% file: data/intro.tex
\section{Introduction}

Dense correspondence estimation has been a core building block for a variety of computer vision tasks~\cite{scharstein2002taxonomy,horn1981determining}. Recently, traditional instance-level correspondence has been extended to the problem setting of semantic-level matching, which aims to align different object instances of the same category~\cite{liu2010sift}. Semantic matching has attracted growing attention~\cite{han2017scnet,Rocco2018,lee2019sfnet} and demonstrated practical value in real-world applications~\cite{taniai2016joint}. However, an effective strategy remains elusive largely due to the presence of background clutters, severe intra-class variation and viewpoint change, as well as difficulty in data annotation.



To tackle those challenges, recent research efforts have adopted the learning-based representation, which have significantly improved semantic matching quality compared with traditional methods~\cite{novotny2017anchornet,kim2019fcss,han2017scnet,Rocco2017,lee2019sfnet}. Early learning methods in semantic matching require strong supervision in the form of ground truth correspondences, which is difficult to obtain for a large set of real images~\cite{ham2017proposal}. As a result, recent trend has focused on weakly-supervised methods~\cite{Rocco2018,huang2019dynamic} that only employ matching image pair, but they are mostly limited to small-scale datasets. A more promising strategy is to leverage self-supervised learning where image pairs and ground-truth correspondences are generated synthetically using random transformations~\cite{Rocco2017,hongsuck2018attentive,lee2019sfnet}. This can significantly reduce annotation cost and makes it possible to utilize large-scale single image datasets along with their existing labels (e.g. masks) for additional constraints~\cite{lee2019sfnet}.  

Despite the increasingly large training datasets, existing deep network based semantic matching typically rely on a deterministic neural networks which do not incorporate uncertainty in their correspondence estimation. Due to the implicit ambiguity in feature matching, such simplistic approaches lack the capacity to measure the quality of prediction results and to properly refine their initial predictions. While some prior work attempt to improve initial estimation through an iterative refinement process~\cite{Rocco2017,hongsuck2018attentive}, they often suffer from error propagation when initial predictions are of low quality.  




In this paper, we propose to incorporate uncertainty reasoning into self-supervised semantic matching in order to estimate and correct low-quality correspondence results. Toward this goal, we first introduce a pixel-wise confidence estimation mechanism to determine whether the initial prediction for each location on image plane is reliable or not. We then develop a confidence-aware refinement procedure to update initial predictions. We exploit the self-supervision setting to generate dense ground truth labels for confidence estimation and introduce an additional adversarial loss to improve overall prediction consistency. A key advantage of our approach is the tight integration of the model design and self-supervised learning strategy, which allows us to effectively train this confidence-aware semantic matching network.    

\begin{figure}[t!]
	\centering
	\includegraphics[width=0.7\linewidth]{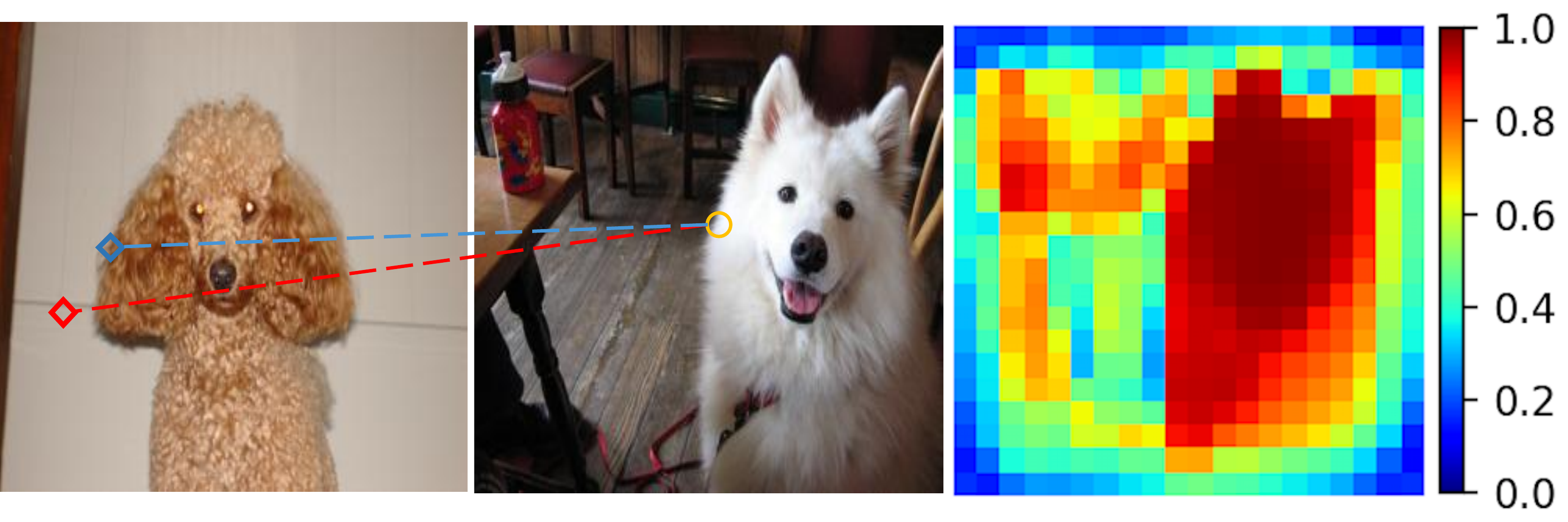}
	\caption{\textbf{Motivation.} From left to right are the source image, the target image, and the estimated confidence map. The point in the target image (yellow circle) was initially matched to the wrong position in the source image (background, red rhombus). After a step of correction through the confidence map, it was matched to a more correct position (foreground, blue diamond).
	}
	\label{fig:add}
\end{figure}



Specifically, we present a novel Confidence-Aware Semantic Matching Network~(CAMNet) that estimates matching confidences and base on which refines the correspondence prediction. Our network consists of four parts. First, a Base Correspondence Network produces initial correspondence predictions. Then, a Confidence Estimation Network is developed to measure pixel-wise confidence of initial predictions. Moreover, taking as input initial predictions and estimated confidence maps, we design a Confidence-aware Refinement Network to produce refined correspondences by propagating reliable predictions. 
We train our model with an alignment loss, a confidence loss and an adversarial loss in an end-to-end fashion. As we develop all the training signals for correspondence, confidence, and adversarial learning from synthetically warped image pairs, we refer to this novel learning scheme as confidence-aware adversarial learning for self-supervised semantic matching.

We extensively evaluate our method on two standard benchmarks, including PF-Willow~\cite{ham2017proposal} and PF-PASCAL~\cite{ham2017proposal}. Our experimental results demonstrate the strong performance of our model over the prior state-of-the-art approaches. Our main contribution are threefold:


\begin{itemize}

	\item We propose a confidence-aware refinement strategy for semantic matching and are the first to consider generating confidence ground truth with the self-supervised setting. 
		
	\item We introduce adversarial training to mitigate the distortion problem in self-supervised semantic matching task.
	
	\item Our self-supervised learning strategy achieves top performance in two standard benchmarks.
\end{itemize}

%% file: data/related_work.tex
\section{Related Work}

\subsection{Semantic Correspondence}
Early semantic matching approaches usually leverage ground truth correspondences for strong supervision~\cite{kim2019fcss,ham2017proposal}. Collecting such annotations under large intra-class appearance and shape variations is label-intensive and may be objective. Consequently, recent work has focused on weakly-supervised~\cite{Rocco2018,huang2019dynamic} or self-supervised setting. Self supervision has attracted growing attention as it further saves human effort and enables to leverage large-scale single image datasets along with their existing labels for additional constraints. Rocco et al.~\cite{Rocco2017} first propose to synthetically generate the image pair and ground truth correspondences from an image itself. Seo et al.~\cite{hongsuck2018attentive} extend this idea with an offset-aware correlation kernel to filter out distractions. Junghyup et al.~\cite{lee2019sfnet} further utilize images annotated with binary foreground masks and subject to synthetic geometric deformations for this task. However, few methods consider the uncertainty in model prediction or incorporate an estimated prediction quality in a refinement process in order to correct partial errors, especially in challenging cases.

\subsection{Confidence Estimation}

Confidence estimation has been widely explored and showing promissing improvement in stereo matching and depth completion~\cite{gidaris2017detect,kim2018unified}. Recently, Sangryul et al. first introduce confidence in weakly-supervised semantic matching~\cite{jeon2019joint}. However, confidence is learned without any available supervision and is used for regularization in loss function instead of reasoning during prediction. In contrast, we are the first to consider generating confidence ground truth as supervision, and directly utilize confidence in model design to guide refined correspondence prediction in self-supervised semantic matching.

\subsection{Generative Adversarial Networks}

Generative adversarial networks (GANs) have gained increasing attention~\cite{bao2017cvae,iizuka2017globally,perarnau2016invertible}. Inspired from the success of GANs, we introduce adversarial learning to enforce the global consistency of warped images, which has never been explored before in semantic matching. We utilize a discriminator to distinguish real data from fake data, and simultaneously to guide the generator to produce higher quality correspondences. As the real data for training GANs are from synthetically warped image pairs, we call our learning scheme as self-supervised adversarial training.


%% file: data/method.tex
\section{Model Architecture}
\label{sec:method}


\subsection{Overview}\label{subsec:overview}
Semantic alignment aims to establish dense correspondences between a source image $\mathbf{I}_{s}$ and a target image $\mathbf{I}_{t}$. A typical CNN-based method computes a correlation map between the convolution features of two images, from which a dense flow field is predicted as final output. 

Our objective is to augment the correspondence prediction with a confidence estimation mechanism so that the model is capable of propagating reliable information during inference and producing a consistent flow prediction from the correlation map.
To achieve this, we adopt a refinement strategy: given an initial dense correspondence between two feature maps, we first estimate the probability of being correctly predicted for each feature location. The estimated confidence map is then fed into a refinement process to guide information propagation and prediction of final correspondence. 

\begin{figure}[t]
	\centering
	\includegraphics[width=1.0\linewidth]{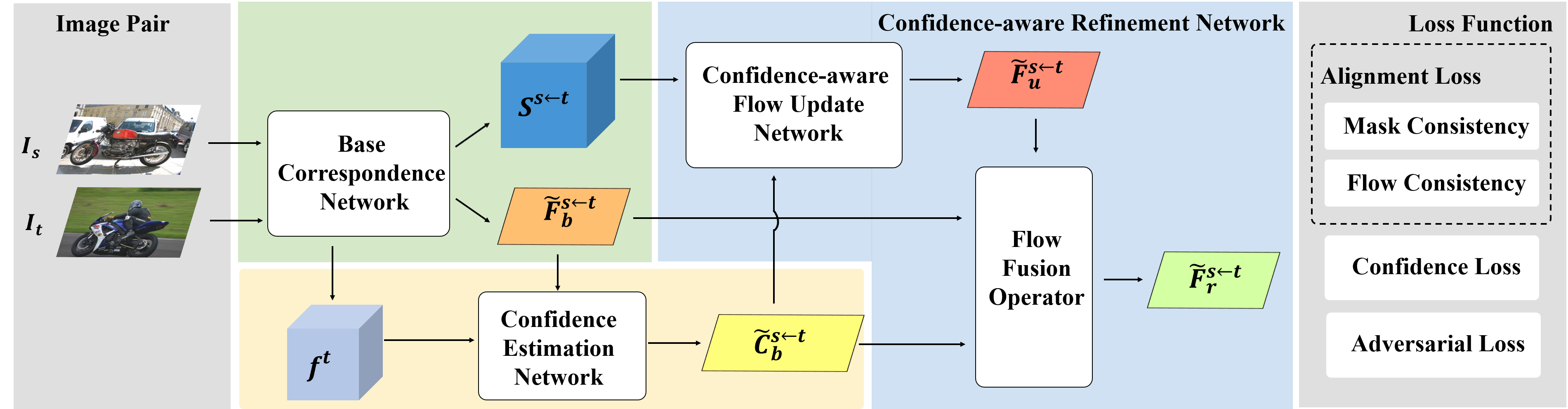}
	\caption{\textbf{Overview of the Confidence-aware Semantic Matching Network.}}
	\label{fig:overview}
\end{figure}

In this section, we introduce a novel network dubbed Confidence-aware Semantic Matching Network (CAMNet) to implement our strategy.
Our network comprises three main submodules, including a Base Correspondence Network (Sec.~\ref{subsec:basenet}), a Confidence Estimation Network (Sec.~\ref{subsec:confi}), and a Confidence-aware Refinement Network (Sec.~\ref{subsec:refinenet}). The entire network is learned in a jointly manner. Below we will describe these submodules of CAMNet in details, and defer the discussion of model learning to Sec.~\ref{subsec:loss}. An overview of our model architecture is shown in Figure~\ref{fig:overview}. 

\subsection{Base Correspondence Network}\label{subsec:basenet}



The first module of our CAMNet is the Base Correspondence Network, aiming to produce an initial semantic flow given the input image pair. To this end, we adopt a differentiable semantic correspondence network as our module design, which consists of a feature extractor, a correlation computation operator and a base flow prediction module\footnote{We note that any differentiable semantic correspondence network can be used here.}.  

Concretely, the input image pair $\mathbf{I}_{s}$ and $\mathbf{I}_{t}$ are first passed through a CNN-based feature extractor, which extract features maps ${f}^{s} \in\mathbb{R}^{d\times h_s\times w_s}$ and ${f}^{t} \in\mathbb{R}^{d\times h_t\times w_t}$  respectively. We adopt ResNet-101~\cite{he2016deep} with additional adaptation layers to parameterize feature extractors as in~\cite{lee2019sfnet}. Subsequently, we compute a 3D correlation map $S^{s \leftarrow t} \in\mathbb{R}^{(h_{s} \times w_{s}) \times h_{t} \times w_{t}}$ from L2 normalized feature map ${f}^{s}$ and ${f}^{t}$, which describes pairwise similarity for any two locations between $\mathbf{I}_{s}$ and $\mathbf{I}_{t}$. 
Given 3D correlation map $S^{s \leftarrow t}$, we then apply a flow prediction module to obtain initial semantic flow predictions $\tilde{\mathcal{F}}^{s \leftarrow t}_b \in\mathbb{R}^{2\times h_t \times w_t}$. While any differentiable flow prediction networks can be adopted here, we choose the kernel soft argmax layer proposed in~\cite{lee2019sfnet} due to its superior performance. The initial semantic flow $\tilde{\mathcal{F}}^{s \leftarrow t}_b$ will be refined in the later process.

\subsection{Confidence Estimation Network}\label{subsec:confi}
To correct partial errors in the initial prediction, we introduce a confidence estimation mechanism to determine what information is trustworthy for guiding the refinement. To this end, we design a confidence estimation network to generate a prediction confidence map from the input image features and predicted flows. By taking the input and output into consideration, the network aims to detect inconsistent flow configurations, which may not be easily discovered by checking output predictions alone. 

Concretely, given the target feature map $f^t$ and the initial flow $\tilde{\mathcal{F}}^{s \leftarrow t}_b$, the confidence estimation network $\mathbb{C}$ estimates a base confidence map $\tilde{\mathcal{C}}^{s \leftarrow t}_b  \in\mathbb{R}^{1\times h_{t}\times w_{t}}$ as follows:
\begin{equation}
\tilde{\mathcal{C}}^{s \leftarrow t}_b = \mathbb{C}(f^{t}, \tilde{\mathcal{F}}^{s \leftarrow t}_b; \theta_ \mathbb{C})
\end{equation} where $\theta_ \mathbb{C}$ is learnable parameter. $\mathbb{C}(\cdot)$ is implemented by fully convolutional layers followed by a softmax operator to normalize score at each location to between 0 to 1, where higher score indicates higher confidence of being correct. The estimated confidence map $\tilde{\mathcal{C}}^{s \leftarrow t}_b$ will serve as an informative guidance in the later refinement process.

\subsection{Confidence-aware Refinement Network}\label{subsec:refinenet}
Our third module, Confidence-aware Refinement Network, aims at refining the initial flow under the guidance of estimated confidence map. It consists of a confidence-aware flow update and a flow fusion operator described as follows.

\subsubsection{Confidence-aware Flow Update}
Different from the initial flow prediction network (cf. Sec.~\ref{subsec:basenet}), our confidence-aware flow update network takes both the confidence map and the correlation map as input, and aims to propagate reliable matching information for flow update. 

Specifically, we feed the confidence map $\tilde{\mathcal{C}}^{s \leftarrow t}_b$ for the initial flow and the correlation map $S^{s \leftarrow t}$ into the confidence-aware flow update network $\mathbb{U}(\cdot)$ and compute an updated flow $\tilde{\mathcal{F}}^{s \leftarrow t}_u \in\mathbb{R}^{2\times h_{t} \times w_{t}}$ as follows:
\begin{equation}
\tilde{\mathcal{F}}^{s \leftarrow t}_u = \mathbb{U}(\tilde{\mathcal{C}}^{s \leftarrow t}_b, S^{s \leftarrow t}; \theta_\mathbb{U}) 
\end{equation} where $\theta_\mathbb{U}$ is learned parameter.

\subsubsection{Flow Fusion}
Finally, we feed the base flow, updated flow and the base confidence map into a flow fusion operator to obtain the final refined flow $\tilde{\mathcal{F}}^{s \leftarrow t}_r \in\mathbb{R}^{2\times h_{t} \times w_{t}}$, which enable us to explicitly suppress  initial predictions with low-confidence. The confidence maps act as gates that control which pixel position initial predictions $\tilde{\mathcal{F}}^{s \leftarrow t}_b$ will be replaced by the updated flow $\tilde{\mathcal{F}}^{s \leftarrow t}_u$. We choose the following flow fusion operator to generate high quality refined flows:
\begin{equation}
\tilde{\mathcal{F}}^{s \leftarrow t}_r =  \tilde{\mathcal{F}}^{s \leftarrow t}_b  \odot \tilde{\mathcal{C}}^{s \leftarrow t}_b +  \tilde{\mathcal{F}}^{s \leftarrow t}_u  \odot(1 - \tilde{\mathcal{C}}^{s \leftarrow t}_b).
\end{equation}
where $\odot$ denotes the point-wise product of confidence maps and flows.

%% file: data/loss.tex
\section{Confidence-aware Adversarial learning}\label{subsec:loss}


We utilize a self-supervised learning strategy for network training by exploiting an image dataset with foreground segmentation masks. Our method first transforms each image and its corresponding foreground mask with random transformations to generate training image pairs as in~\cite{lee2019sfnet}. During training, we make predictions from target to source and source to target in two directions for each input image pair. 
Based on this setup, we develop a multi-task loss that includes three components: an alignment loss that enforces the bidirectional matching consistency, a confidence loss that supervises the confidence network, and an adversarial loss on the transformed images within foreground region. We will describe these loss terms in detail below and omit multiply every loss with foreground masks for notation brevity.

\subsection{Alignment Loss}

We first adopt the self-supervised training loss proposed in SFNet~\cite{lee2019sfnet} to encourage correspondences to be established within foreground masks and measure consistency between flow estimations in two directions, which has the following form:
\begin{equation}
\mathcal{L}_a(\tilde{\mathcal{F}}^{s \leftarrow t},\tilde{\mathcal{F}}^{t \leftarrow s}) = \lambda \cdot \mathcal{L}^{mask}(\tilde{\mathcal{F}}^{s \leftarrow t},\tilde{\mathcal{F}}^{t \leftarrow s},M^t,M^s)+\mathcal{L}^{flow}(\tilde{\mathcal{F}}^{s \leftarrow t},\tilde{\mathcal{F}}^{t \leftarrow s})
\end{equation}
where $\lambda$ is the weight parameter for balancing mask and flow consistency loss, $M^t$ and $M^s$ are the binary ground truth foreground masks for the target and the source image respectively. Our alignment loss is defined on the initial flow and the refined flow jointly:
\begin{equation}
\mathcal{L}_{align} = \gamma \cdot \mathcal{L}_a(\tilde{\mathcal{F}}^{s \leftarrow t}_b,\tilde{\mathcal{F}}^{t \leftarrow s}_b)+  \mathcal{L}_a(\tilde{\mathcal{F}}^{s \leftarrow t}_r,\tilde{\mathcal{F}}^{t \leftarrow s}_r)
\end{equation} where $\gamma$ is used to balance loss for base flow and refined flow.

\subsection{Confidence Loss}\label{subsubsec:confiloss}
%
%
%
%
%

As we have ground truth flow in self-supervised learning, the ground truth base confidence map $\mathcal{C}^{s \leftarrow t}_b$ can be obtained by thresholding the error between the predicted base flow $\tilde{\mathcal{F}}^{s \leftarrow t}_b$ and the ground truth flow $\mathcal{F}^{s \leftarrow t}$. This enables us to directly supervise confidence estimation. The confidence loss measures the quality between estimated confidence map and the ground truth confidence map as follows:
\begin{equation}
\mathcal{L}_{c}(\tilde{\mathcal{C}}^{s \leftarrow t},\tilde{\mathcal{C}}^{t \leftarrow s},\mathcal{C}^{s \leftarrow t},\mathcal{C}^{t \leftarrow s}) = \mathbb{CE}(\tilde{\mathcal{C}}^{s \leftarrow t},\mathcal{C}^{s \leftarrow t}) + \mathbb{CE}(\tilde{\mathcal{C}}^{t \leftarrow s},\mathcal{C}^{t \leftarrow s})
\end{equation}where $\mathbb{CE}$ is the cross entropy loss. 

In addition to the confidence maps $\tilde{\mathcal{C}}^{t \leftarrow s}_b$ and $\tilde{\mathcal{C}}^{t \leftarrow s}_b$, we also estimate refined confidence maps $\tilde{\mathcal{C}}^{t \leftarrow s}_r$ and $\tilde{\mathcal{C}}^{s \leftarrow t}_r$ for the refined flow in order to further regularize confidence estimation network. Hence our final confidence loss is defined as 
\begin{equation}
	\mathcal{L}_{confi} =  \beta \cdot  \mathcal{L}_{c}(\tilde{\mathcal{C}}^{s \leftarrow t}_b,\tilde{\mathcal{C}}^{t \leftarrow s}_b,\mathcal{C}^{s \leftarrow t}_b,\mathcal{C}^{t \leftarrow s}_b)+
	\mathcal{L}_{c}(\tilde{\mathcal{C}}^{s \leftarrow t}_r,\tilde{\mathcal{C}}^{t \leftarrow s}_r,\mathcal{C}^{s \leftarrow t}_r,\mathcal{C}^{t \leftarrow s}_r)
\end{equation}
where $\beta$ is the weight parameter balancing the two loss terms.

\begin{figure}[t]
	\centering
	\includegraphics[width=0.55\linewidth]{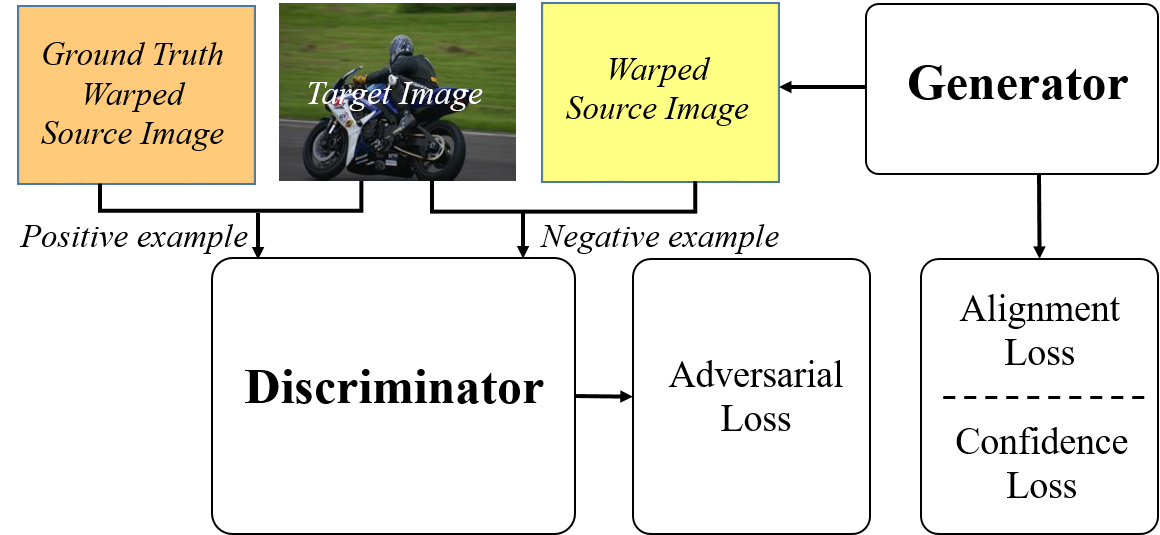}
	
	\caption{\textbf{Overview of self-supervised adversarial training for semantic matching.} The negative example and positive example composed of the target image and the warped source image are important for adversarial training. The discriminator and the generator are trained alternately.}
	\label{fig:ganloss}
\end{figure}

\subsection{Adversarial Loss}\label{subsubsec:ganloss}
We introduce an adversarial loss for enforcing global consistency in flow prediction. To this end, we first build an image generator $\mathbf{G}$ by integrating our model (described in Sec.~\ref{sec:method}) with a warping operator $\mathcal{W}$. Considering matching from target to source, $\mathbf{G}$ generates synthetically warped source images $\tilde{\mathbf{I}}_{t}$ given an image pair $( \mathbf{I}_s,\mathbf{I}_t)$ as follows:
\begin{equation}
\tilde{\mathbf{I}}_{t}=\mathcal{W}\left(\mathbf{I}_{s} ; \tilde{\mathcal{F}}^{s \leftarrow t}_r\right)=\mathbf{G}(\mathbf{I}_s,\mathbf{I}_t)
\end{equation}

We define a Discriminator $\mathbf{D}$ to distinguish warped images from the ground-truth flows and the generated flows. The input of the discriminator contains either source images warped by the ground truth flow (denoted as $\mathbf{I}_t^*$) or by the predicted flow, i.e. $\tilde{\mathbf{I}}_t$, and both of them are concatenated with the corresponding target image. From the input pair, the discriminator should learn whether the flow is correct s.t. the warped source image align with target image. $\mathbf{D}$ outputs a one channel prediction map with values between 0 to 1 where higher value indicates the more $\mathbf{D}$ believes that the input pair is real data.

In self-supervision, we apply random transformation in the form of ground truth flow $\mathcal{F}^{s \leftarrow t}$ to the single source image $\mathbf{I}_s$ to form a training image pair $( \mathbf{I}_s,\mathbf{I}_t)$. Thus, the source image warped by ground truth flow $\mathbf{I}_t^*$ is essentially the target image $\mathbf{I}_t$ itself in the self-supervised setting:
\begin{equation}
\mathbf{I}_t^* = \mathbf{I}_{t}=\mathcal{W}\left(\mathbf{I}_{s} ; \mathcal{F}^{s \leftarrow t}\right)
\label{eq:training_pair}
\end{equation}

Consequently, we call our framework as self-supervised adversarial learning as real input pair for training GANs is developed from self-supervision (i.e., no real annotations are needed).

Adversarial loss is used for mitigating distortion problem in semantic matching. The discriminator and the generator is trained alternatively as shown in Fig.~\ref{fig:ganloss}. After training is done, the generator (without warping operator) is used as a correspondence estimator and the discriminator can be removed.

\subsubsection{Training the Generator}

The generator $\mathbf{G}$ tries to generate realistic images to fool the discriminator, and our loss term aims to minimize the following Least Square GAN loss bidirectionally
\begin{equation}
\mathcal{L}_{adv} = (\mathbf{D}(\tilde{\mathbf{I}}_s,\mathbf{I}_s)-1)^2+(\mathbf{D}(\tilde{\mathbf{I}}_t,\mathbf{I}_t)-1)^2
\end{equation}
which enforces the generator to produce reasonable warped images.


Our overall loss for the generator consists of the alignment loss, the confidence loss and the adversarial loss terms with  $\mu_1$, $\mu_2$ as weight parameters as follows,

\begin{equation}
\mathcal{L}_{G} =\mu_1 \cdot \mathcal{L}_{align} + \mu_2 \cdot \mathcal{L}_{confi} + \mathcal{L}_{adv}
\end{equation} 



\subsubsection{Training the Discriminator}
The discriminator is used to distinguish predicted flow-image pair from real flow-image pair. Considering both matching directions, the discriminator loss for the real data pair $\mathcal{L}_{real}$ and predicted data pair $\mathcal{L}_{fake}$ is as follows

\begin{equation}
\mathcal{L}_{real} = (\mathbf{D}(\mathbf{I}_s^*,\mathbf{I}_s)-1)^2 + (\mathbf{D}(\mathbf{I}_t^*,\mathbf{I}_t)-1)^2
\end{equation} 
\begin{equation}
\mathcal{L}_{fake} = (\mathbf{D}(\tilde{\mathbf{I}}_s,\mathbf{I}_s)-0)^2 + (\mathbf{D}(\tilde{\mathbf{I}}_t,\mathbf{I}_t)-0)^2
\end{equation}
where $\mathbf{I}_t^*$ is the same as $\mathbf{I}_t$ in self-supervision as shown in Eq.~\ref{eq:training_pair}. Similarly, $\mathbf{I}_s^*$ equals to $\mathbf{I}_s$. Consequently, the overall loss for the discriminator is defined as:
\begin{equation}
\mathcal{L}_{D} = \mathcal{L}_{real} + \mathcal{L}_{fake}
\end{equation}

%% file: data/experiments.tex
\section{Experiments}

We conduct experiments on standard datasets PF-PASCAL~\cite{ham2017proposal} and PF-WILLOW~\cite{ham2017proposal} to evaluate our method for semantic matching. We first present the implementation details in Sec.\ref{sec:implementation}. Then, we show quantitative and qualitative results of the two datasets in Sec.\ref{sec:pf-pascal-benchmark} and Sec.\ref{sec:pf-willow-benchmark}. Ablation study is given in Sec.\ref{sec:ablation-study}. 

\subsection{Implementation details}
\label{sec:implementation}
Images are resized into the size of 320$\times$320. Following SFNet~\cite{lee2019sfnet}, our model is trained on the PASCAL VOC 2012 segmentation dataset~\cite{everingham2010pascal} which includes a foreground mask for each image and PF-PASCAL~\cite{ham2017proposal} validation split is used for model selection. The confidence estimation network is implemented with 3$\times$3 convolutional filters sequentially followed by BN and ReLU. We multiply correlation map by estimated confidence map as the input of the confidence-aware refinement network. The confidence-aware refinement network is implemented with architecture similar to Dense-Net and the last layer of which is of two channels. We adopt PatchGAN~\cite{isola2017image} as our discriminator. To validate generalization ability of our method, we test our trained model on PF-WILLOW dataset~\cite{ham2017proposal} test split without finetuning. We set $\lambda=0.188$, $\gamma=\beta=0.4$, $\mu_1=288.0$, $\mu_2=18.0$.



\begin{figure}[t]
	\centering
	\includegraphics[width=0.8\linewidth]{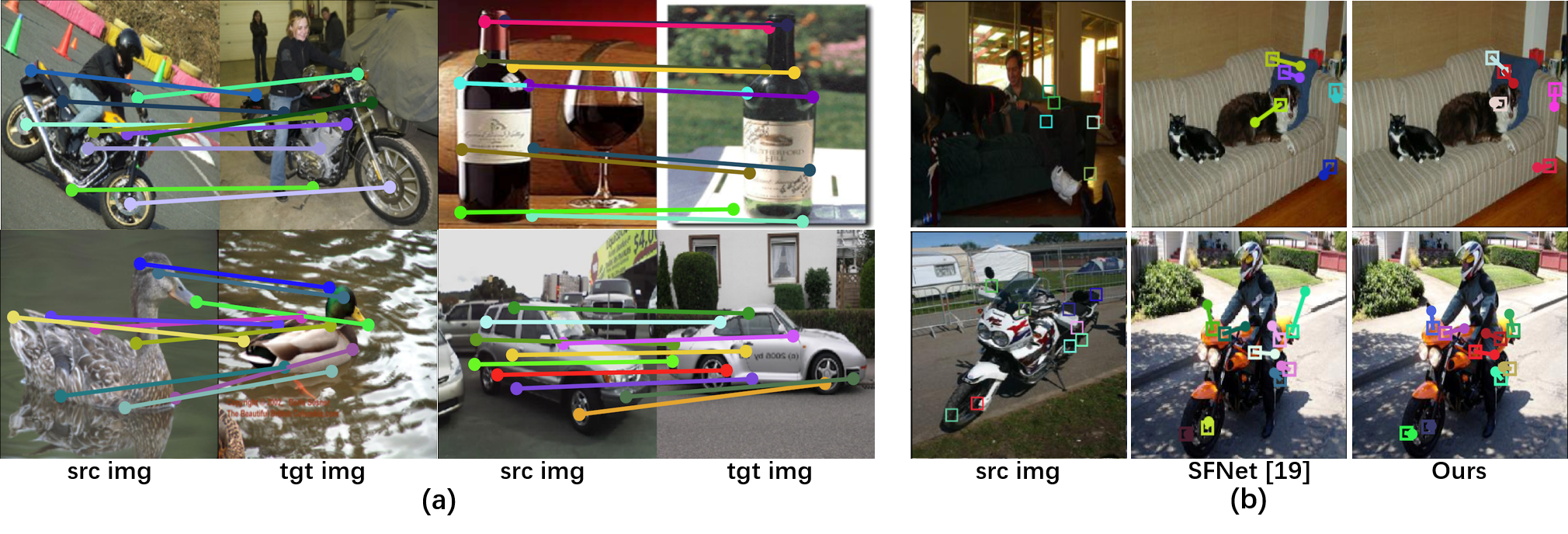}
	\vspace{-4mm}

	\caption{(a) Alignment examples on PF-WILLOW from our model. (b) Qualitative comparisons on PF-PASCAL. We show the ground truth and predicted keypoints in squares and dots respectively, with their distance in target images depicting the matching error.}
	\label{fig:vis_willow_pascal}
	
\end{figure}

\subsection{PF-WILLOW Benchmark}
\label{sec:pf-willow-benchmark}
\subsubsection{Dataset and evaluation metric}The PF-WILLOW dataset \cite{ham2017proposal} contains 900 image pairs selected from 100 images with four categories. We report the PCK scores~\cite{yang2012articulated} $(\alpha = 0.05, 0.10, 0.15)$ w.r.t bounding box size.

\begin{table}[t]
	\centering
	\caption{\textbf{Evaluation results on PF-WILLOW~\cite{ham2017proposal} test split.}}
	\vspace{-1mm}
	\resizebox{0.8\textwidth}{!}{
		\begin{tabular}{c|c|c|c|c}
			\hline
			Method    & Supervision&\begin{tabular}[c]{@{}c@{}}PCK($\alpha=0.05$)\end{tabular} & \begin{tabular}[c]{@{}c@{}}PCK($\alpha=0.10$)\end{tabular} & \begin{tabular}[c]{@{}c@{}}PCK($\alpha=0.15$)\end{tabular} \\ \hline\hline
			\begin{tabular}[c]{@{}c@{}}HOG+PF-LOM \cite{ham2017proposal}\end{tabular} &-&0.284&0.568&0.682  \\ \hline
			\begin{tabular}[c]{@{}c@{}}DCTM \cite{kim2017dctm}\end{tabular}   &Strong&0.381&0.610&0.721\\
			\begin{tabular}[c]{@{}c@{}}UCN-ST \cite{choy2016universal}\end{tabular}  &Strong &0.241&0.540&0.665\\
			\begin{tabular}[c]{@{}c@{}}CAT-FCSS \cite{kim2019fcss}\end{tabular} &Strong&0.362&0.546&0.692\\
			\begin{tabular}[c]{@{}c@{}}SCNet
				~\cite{han2017scnet}\end{tabular}  &Strong &0.386&0.704&0.853\\ \hline
			\begin{tabular}[c]{@{}c@{}}WeakAlign
				~\cite{Rocco2018}\end{tabular}&Weak & 0.382                                                   & 0.712                                                   & 0.858                                                   \\ %
			\begin{tabular}[c]{@{}c@{}}RTN \cite{kim2018recurrent}\end{tabular} &Weak      & 0.413                                                   & 0.719                                                   & 0.862                                                   \\
			\begin{tabular}[c]{@{}c@{}}NCNet \cite{Rocco18b}\end{tabular}  &Weak   & 0.440                                                   & 0.727                                                   & 0.854                                                   \\  \hline\hline
			\begin{tabular}[c]{@{}c@{}}CNNGeo~\cite{Rocco2017}\end{tabular}&Self&-&0.560&- \\
			\begin{tabular}[c]{@{}c@{}}A2Net~\cite{hongsuck2018attentive}\end{tabular}&Self&-&0.680&- \\
			\begin{tabular}[c]{@{}c@{}}SFNet~\cite{lee2019sfnet}\end{tabular} &Self (+mask)     & 0.459                                                   & 0.735                                                   & 0.855                                                   \\ 
			Ours   &Self (+mask)   &                     \textbf{0.464}                               &                            \textbf{0.746}                     &                            \textbf{0.863}                      \\ \hline
	\end{tabular}}
	
	\label{tab:evalwillow}
\end{table}

\subsubsection{Experimental Results} Table~\ref{tab:evalwillow} compares our CAMNet with other SOTA approaches. The PCK values ($\alpha = 0.05, 0.10, 0.15$) of our method are $46.4\%$, $74.6\%$ and $86.3\%$ respectively, outperforming the previously published best self-supervised method SFNet by $0.5\%$, $1.1\%$, and $0.8\%$ respectively. Fig.~\ref{fig:vis_willow_pascal} shows that our method can effectively handle background clutters and viewpoint changes.

\subsection{PF-PASCAL Benchmark}
\label{sec:pf-pascal-benchmark}

\begin{table}[t]
	\centering
	\caption{\textbf{Evaluation results on PF-PASCAL~\cite{ham2017proposal} test split.}}
	\resizebox{0.8\textwidth}{!}{
		\begin{tabular}{c|c|c|c|c}
			\hline
			Method   &Supervision & \begin{tabular}[c]{@{}c@{}}PCK($\alpha= 0.05$)\end{tabular} & \begin{tabular}[c]{@{}c@{}}PCK($\alpha= 0.10$)\end{tabular} & \begin{tabular}[c]{@{}c@{}}PCK($\alpha= 0.15$)\end{tabular} \\ \hline
			\hline
			\begin{tabular}[c]{@{}c@{}}ProposalFlow \cite{ham2017proposal}\end{tabular} &-  &0.314&0.625&0.795\\
			\begin{tabular}[c]{@{}c@{}}DCTM~\cite{kim2017dctm}\end{tabular}   &Strong&0.342&0.696&0.802\\
			\begin{tabular}[c]{@{}c@{}}SCNet~\cite{han2017scnet}\end{tabular}   &Strong&0.362&0.722&0.820\\ \hline
			\begin{tabular}[c]{@{}c@{}}WeakAlign~\cite{Rocco2018}\end{tabular} &Weak& 0.460                                                   & 0.758                                                   & 0.884                                                   \\
			\begin{tabular}[c]{@{}c@{}}RTN \cite{kim2018recurrent}\end{tabular}       &Weak& 0.552                                                   & 0.759                                                   & 0.852                                                   \\
			\begin{tabular}[c]{@{}c@{}}NC-Net \cite{Rocco18b}\end{tabular}           &Weak& 0.543                                                   & 0.789                                                   & 0.860                                                   \\
			\begin{tabular}[c]{@{}c@{}}SAM-Net ~\cite{kim2019semantic}\end{tabular}          &Weak& \textbf{0.601}                                                  & 0.802                                                   & 0.869                                                   \\\hline
				\begin{tabular}[c]{@{}c@{}}CNNGeo~\cite{Rocco2017}\end{tabular}&Self&-&0.600&- \\
			\begin{tabular}[c]{@{}c@{}}A2Net~\cite{hongsuck2018attentive}\end{tabular}&Self&-&0.680&- \\
			\begin{tabular}[c]{@{}c@{}}SFNet~\cite{lee2019sfnet}\end{tabular}     &Self (+mask)&0.536                                                   & 0.819                                                   & 0.906                                                   \\ 
			Ours    &Self (+mask) &        0.549                               &                   \textbf{0.835}                           &            \textbf{0.910}                    \\ \hline
	\end{tabular}}
	
	\label{tab:evalpascal}
\end{table}

\subsubsection{Dataset and evaluation metric}

There are 1351 image pairs on PF-PASCAL \cite{ham2017proposal} benchmark. The key point annotations are only used for evaluation. In line with previous works, we report PCK~\cite{yang2012articulated} ($\alpha = 0.05, 0.10, 0.15$) w.r.t image size.

\vspace{-3mm}
\subsubsection{Experimental Results} 

Table \ref{tab:evalpascal} shows the detailed comparison. Our method achieves the best results on $\alpha=0.05, 0.1, 0.15$ in self-supervised setting, which outperforms the previous best self-supervised method SFNet~\cite{lee2019sfnet} by $1.3\%$, $1.6\%$ and $0.4\%$ respectively, demonstrating our method's effectiveness. Fig.~\ref{fig:vis_willow_pascal} shows qualitative comparision between our model and SFNet~\cite{lee2019sfnet}.

%

\subsection{Ablation Study}
\label{sec:ablation-study} 

We select SFNet \cite{lee2019sfnet} as our baseline and report PCK($\alpha=0.1$) on PF-PASCAL~\cite{ham2017proposal} and PF-WILLOW~\cite{ham2017proposal} test split to analyse the effectiveness of our proposed individual modules.
As shown in Table~\ref{tab:ablation}, PCK results increase steadily when sequentially adding our proposed modules, leading to $1.1\%$ and $1.6\%$ improvement on the PF-WILLOW and PF-Pascal respectively in the end. Although adding GANs does not show significant improvements in numbers, it indeed mitigates distortion problems and provides better visual quality, and does not add any additional computational complexity during inference.
\begin{table}[t]
	\centering
	\caption{\textbf{Ablation study on PF-PASCAL~\cite{ham2017proposal} and PF-WILLOW~\cite{ham2017proposal}.}}
	\resizebox{0.8\textwidth}{!}{
		\begin{tabular}{c|c|c}
			\hline
			Method    &

			 \begin{tabular}[c]{@{}c@{}} PCK (PF-WILLOW)\end{tabular} & \begin{tabular}[c]{@{}c@{}}PCK (PF-Pascal)\end{tabular} \\ 
			 
			\hline
			\hline
			\begin{tabular}[c]{@{}c@{}}SFNet \cite{lee2019sfnet}\end{tabular}         & 0.735                                                   & 0.819                                \\ \hline
			Baseline+Refinement& 0.740 & 0.825                                \\ 
			Baseline+Confidence-aware Refinement& 0.745 & 0.833                                \\ 
			Baseline+Confidence-aware Refinement+GAN&  \textbf{0.746}  &   \textbf{0.835}                         \\ \hline
	\end{tabular}}
	\label{tab:ablation}
\end{table}

%% file: data/conclusion.tex
\section{Conclusion}

In this paper, we proposed an effective deep network CAMNet for semantic matching. First, we developed a confidence-aware refinement procedure by directly utilizing confidence in model design to guide refined correspondence prediction. In addition, we introduce adversarial training to mitigate distortion in semantic alignment. Finally, we design a novel end-to-end framework with self-supervision to enable effective confidence and adversarial learning. Experimental results on two standard benchmarks confirmed the effectiveness of our method.